\documentclass[ 
]{ceurart}

\usepackage{listings}
\usepackage{url}
\usepackage{hyperref}
\usepackage{subcaption}
\usepackage{float}

\begin{document}

\copyrightyear{2026}
\copyrightclause{Copyright for this paper by its authors.
  Use permitted under Creative Commons License Attribution 4.0
  International (CC BY 4.0).}

\conference{Data-Driven Storytelling: Bridging Knowledge Graphs, GenAI, and Narrative (DDS2026), in ISWC 2026 Workshops Joint Proceedings, October 25--26, 2026, Bari, Italy}

\title{From Queries to Narratives: Cultural Heritage Data Stories for Knowledge Graph Exploration and Quality Assessment}


\author[1,2]{Tabea Tietz}[%
orcid=0000-0002-1648-1684,
email=tabea.tietz@fiz-karlsruhe.de,
]
\author[3]{Torsten Schrade}[%
orcid=0000-0002-0953-2818,
email=torsten.schrade@adwmainz.de,
]
\author[1]{Etienne Posthumus}[%
orcid=0000-0002-0006-7542,
email=etienne.posthumus@partners.fiz-karlsruhe.de,
]
\author[3]{Linnaea Söhn}[%
orcid=0000-0001-8341-1187,
email=linnaea.soehn@adwmainz.de,
]
\author[3]{Jonatan Jalle Steller}[%
orcid=0000-0002-5101-5275,
email=jonatan.steller@adwmainz.de,
]
\author[1,2]{Jörg Waitelonis}[%
orcid=0000-0001-7192-7143,
email=joerg.waitelonis@fiz-karlsruhe.de,
]
\author[1,2]{Harald Sack}[%
orcid=0000-0001-7069-9804,
email=harald.sack@fiz-karlsruhe.de,
]
\fntext[1]{Contributor Roles: Tabea Tietz: Conceptualization, Validation, Writing – original draft; Torsten Schrade: Software, Writing – review \& editing, Funding acquisition; Etienne Posthumus: Software, Resources, Data curation, Validation, Writing - review and editing; Linnaea Söhn: Conceptualization, Software, Resources, Data Curation, Validation, Writing - review and editing; Jonatan Jalle Steller: Software, Validation; Jörg Waitelonis: Supervision; Harald Sack: Writing – review \& editing; Supervision; Project administration; Funding Acquisition}

\address[1]{FIZ Karlsruhe – Leibniz Institute for Information Infrastructure, Eggenstein-Leopoldshafen, Germany}
\address[2]{Institute of Applied Informatics and Formal Description Methods (AIFB) of KIT, Karlsruhe, Germany}
\address[3]{Academy of Sciences and Literature Mainz, Geschwister-Scholl-Straße 2, 55131 Mainz, Germany}

\begin{abstract}
  Cultural-heritage knowledge graphs such as the NFDI4Culture Knowledge Graph contain millions of triples about artworks, music, inscriptions, historical events, and the people and places connected to them. For many users, however, discovering this knowledge can be difficult. While SPARQL can be learned, writing meaningful queries first requires an in-depth understanding of the graph's data model, an investment many domain researchers and practitioners are unwilling to make. Even with existing user interfaces, a starting point and some guidance are usually needed, because the data contained in the graph is highly specialized, heterogeneous, and constantly growing, making it challenging to know what it contains or which questions it can answer. 
  In this paper, we present data stories as a way not only to lower this barrier, but also to turn exploration into data-quality assessment, and thus combine accessible querying with the discovery of issues that remain hidden in aggregate statistics. In this contribution, a data story is understood as a narrative document that integrates explanatory text and images with executable SPARQL queries and their visualized results. It is described how they are authored against the graph and how they serve several purposes: guiding users through an unfamiliar graph, creating reproducible narratives, and surfacing data-quality issues previously hidden in aggregate statistics. The authoring platform LODEON including its Sparnatural and AI-supported authoring assistants is introduced as a proof-of-concept. Within the authoring environment, every claim made about the data can be backed by an explicit query, making these narratives transparent and reproducible. This paper also reflects on lessons learned from hands-on seminars with Master's students and an international workshop held at DH2026. Early experience suggests that such data stories make cultural-heritage knowledge graphs more accessible for both exploration and quality assessment.

\end{abstract}

\begin{keywords}
  Cultural Heritage \sep
  Knowledge Graph \sep
  Storytelling \sep
  LLM \sep
  Data Story 
\end{keywords}

\maketitle

\section{Introduction}
\label{sec:intro}

The NFDI4Culture Knowledge Graph\footnote{\url{https://nfdi4culture.de/services/details/culture-knowledge-graph.html}} (NFDI4Culture-KG) currently contains approximately 150 million triples relating to data on material and immaterial cultural heritage, such as artworks, musical sources, inscriptions, historical events, and the people and places that connect them. Its scale makes it a valuable resource for the (German) cultural-heritage community, but also a challenge to approach for new users, domain experts, and the data providers whose collections it integrates. The graph indexes data from numerous data portals across a wide range of domains ranging from musicology, art history, architecture to the performing arts and media studies, each contributing according to their subject-specific standards. Like many domain-specific KGs, it is highly specialized, heterogeneous, and constantly growing in a way that users struggle to know what it contains or which questions it can help to answer. The graph can be explored through a public SPARQL endpoint. SPARQL can be learned, but writing meaningful queries first requires learning the underlying data model: the Culture Ontology (CTO)~\cite{tietz2025knowledge} at the domain level, the NFDIcore ontology at the mid-level~\cite{waitelonis2025nfdicore,bruns-nfdicore}, and the Basic Formal Ontology (BFO)~\cite{otte2022bfo} on the upper level~\cite{otte2022bfo}. Many domain researchers and practitioners are hesitant to make this investment. The data can therefore also be explored via the Culture Data Search\footnote{\url{https://nfdi4culture.de/datasearch/explore}}, a multimodal interface for finding research data through a free-text and incipit search, filter options, or image input. However, with data this specialized, users still need a starting point and an idea of which questions the graph can answer. Feedback from the community over several years has made clear that users want guidance on what to ask and how to ask it, and they want searches to be reproducible, so that a query is permanent and can become part of the research record. 

Data stories can provide this guidance. We understand a data story as a narrative document about the data in the KG that integrates explanatory text and images with executable SPARQL queries and their visualized results. In this contribution, data stories serve three purposes. First, they onboard new users, who can read a story and get an idea of what the graph contains and how the data in it is connected. Second, data stories allow users to go through an authoring process in which they can begin to explore their own questions. Third, the act of authoring a story surfaces data-quality issues that remain hidden in aggregate statistics.

In this paper we describe how a first NFDI4Culture data-stories environment was implemented based on the graph and realized with SHMARQL\footnote{\url{https://shmarql.com/}}, how it was used in teaching at Karlsruhe Institute of Technology and international workshops, and how the friction observed there motivated an effort towards an AI-assisted authoring and exploration environment. On that basis we introduce LODEON, together with the Sparnatural query builder~\cite{francart2023sparnatural} and an AI authoring assistant, as a proof-of-concept. 

The use cases described in this contribution are tied to the cultural-heritage domain, with the NFDI4Culture-KG and in parts the German Memory Atlas (GeMeA), a KG over approximately 26.8 million objects from the German Digital Library (DDB)\footnote{\url{https://github.com/ISE-FIZKarlsruhe/gemea}}. However, LODEON can serve any KG, and the problems of access, query formulation, and reproducibility it addresses are not specific to this field. 

This paper is structured as follows: Chapter~\ref{sec:related} introduces existing data stories environments and means of exploration. Chapter~\ref{sec:culture} introduces shortly the NFDI4Culture-KG. Chapter~\ref{sec:main} describes the process of developing a data stories environment, discusses lessons learned in teaching and workshops and motivates the need for the AI-assisted environment LODEON. The LODEON prototype is described, followed by a discussion of lessons learned. The conclusion in Chapter~\ref{sec:conclusion} closes the paper.

\section{Related Work}
\label{sec:related}

The approach presented here builds on two lines of research, the narrative presentation of data and the exploration and querying of KGs. Both are discussed in the following.

\subsection{Data Stories and Narrative Visualization}

The idea of guiding an audience through data with a narrative is well known in data journalism~\cite{arrese2022beginning}, where visualized data supports an authored story, for instance in "The Upshot" by The New York Times\footnote{\url{https://www.nytimes.com/international/section/upshot}}. In the Digital Humanities (DH), several tools bring this format closer to research data: the CLARIAH Data Stories Editor offers an interactive environment for creating and publishing data-driven research narratives \cite{sanders2023developing}, and the Carnegie Hall Data Lab experiments present collection data through exploratory, narrative interfaces.\footnote{\url{https://data.carnegiehall.org/datalab/experiments/}} These approaches share the goal of this contribution of making data more approachable through narratives. However, they differ in the way how the narrative relates to the underlying data. A story is typically authored over a fixed export, and a reader cannot see, run, or alter the query that produced a given statement. In the contribution presented in this paper, each claim about the underlying data in a story can be generated by a SPARQL query embedded in the narrative and executed against the live endpoint. As a result, the reader can inspect and re-run the query behind any statement, and a story's results reflect the current state of the data rather than a snapshot fixed at authoring time.

\subsection{Querying and Exploring Knowledge Graphs}

Linked Data browsers such as LodLive~\cite{camarda2012lodlive}, LODmilla~\cite{micsik2015exploring}, and LodView\footnote{\url{https://github.com/linkeddatacenter/app-lodview}} let users navigate entities and their relationships, while query-oriented tools such as YASGUI~\cite{rietveld2013yasgui} and the educational RDF Playground~\cite{inostroza2023rdf} support writing and running SPARQL. Visual query builders like Sparnatural~\cite{francart2023sparnatural} allow users to assemble queries without writing SPARQL by hand. Furthermore, large language models (LLMs) have been used widely to translate natural-language questions into SPARQL~\cite{khorashadizadeh2024research,rony2022sgpt}. Our contribution integrates the capabilities of SHMARQL, a Linked Data publishing platform that enables semantic web professionals to disseminate data, Sparnatural, and an LLM-based "AI Assistant" within a storytelling format that supports the guidance through the data in the KG and ensures reproducibility.

\smallskip

The contributions described above present solutions where narrative visualization offers guidance but not reproducible access to the data or query tools offer access and exploration, but no narrative guidance. The contribution of this paper is to combine both in one storytelling format over a live cultural-heritage KG. 
\section{Bridging the Distance with Data Stories in NFDI4Culture}
\label{sec:culture}

NFDI4Culture is a consortium within the German National Research Data Infrastructure (NFDI) dedicated to material and immaterial cultural heritage. It brings together subject-specific portals and collections from architecture, art history, design, musicology, the performing arts, and media studies, which are maintained by different institutions and described with various metadata standards~\cite{tietz2023damalos,tietz2025knowledge}. To make these decentralized resources findable and interoperable, NFDI4Culture has been creating the NFDI4Culture-KG, a semantic index that offers a single point of access to research data while the data itself remains stored and owned by the contributing institutions. The graph is modeled with the Culture Ontology (CTO), which extends the mid-level ontology NFDIcore and is aligned with BFO~\cite{sack2023cordi}. The NFDI4Culture-KG is published through a public SPARQL endpoint\footnote{\url{https://nfdi4culture.de/resources/knowledge-graph.html}} and can be explored using the graphical user interfaces within the Culture Information Portal.

Data has been integrated into the graph through provider-specific feeds. Once an institution or a researcher is ready to contribute, a data feed is created and maintained via the Culture Information Portal. The provider's data is then processed by an Extract--Transform--Load (ETL) pipeline, called "Culture Kitchen"~\cite{bruns2024kitchen,steller2024DHd}. Currently the KG contains more than 3.7m resources from 18 data feeds contributed by the cultural-heritage community in Germany, with the potential to grow given the 88 data portals listed by NFDI4Culture. Even though aggregating decentralized resources into a shared index is necessary, this alone is not sufficient for their reuse. As Borgman and Groth argue, effective reuse requires overcoming the existing distance between those who create and those who intend to reuse the data~\cite{Borgman2025Distance}.

As part of NFDI4Culture, user stories\footnote{\url{https://nfdi4culture.de/about-us/user-stories.html}} were collected from the German cultural-heritage community to capture requirements for the graph and the Culture Information Portal~\cite{tietz2023damalos}. However, it became clear that many contributors did not have a concrete idea of what the graph would contain, how it should be used, and how it might help to answer specific research questions. Over the years it has become clear that these challenges are not resolved by the detailed ontology documentation, the public SPARQL endpoint, or search interfaces alone, because some conventions that matter for querying only become apparent in use and users require starting points to create meaningful searches using the UI. Therefore it is necessary to explicitly showcase the users what knowledge the graph contains and which questions can be answered. Furthermore, users need to see questions answered against the live graph, i.e. a working query that can be read, adapted to a related question, and cited as a permanent part of the research process. 

In order to fulfill these needs, the data stories laboratory\footnote{\url{https://datastories.nfdi4culture.de}} has been developed within NFDI4Culture featuring an AI-assisted workbench, which is described in the following Chapter~\ref{sec:main}. However, these needs discovered within NFDI4Culture are not specific to this particular project or to cultural heritage, as any KG containing heterogeneous and continuously growing sources confronts its users with some level of distance between a documented schema and the research questions the data can actually answer. Data stories represent a form of knowledge exchange that helps to close that distance.

\section{Data Stories in NFDI4Culture from Consumption to Creation and AI Assisted Exploration}
\label{sec:main}

This section follows data stories in NFDI4Culture from consuming existing stories created by domain experts to the first environment in which users assume the role of creators within a SHMARQL-based authoring environment. This is followed by lessons learned and the introduction of the proof-of-concept workbench LODEON, which adds a visual query builder and an AI assistant for authoring and exploration.

\subsection{Consuming Data Stories as an Entry Point to Structured Data}

\begin{figure}
    \centering
    \includegraphics[width=1\linewidth]{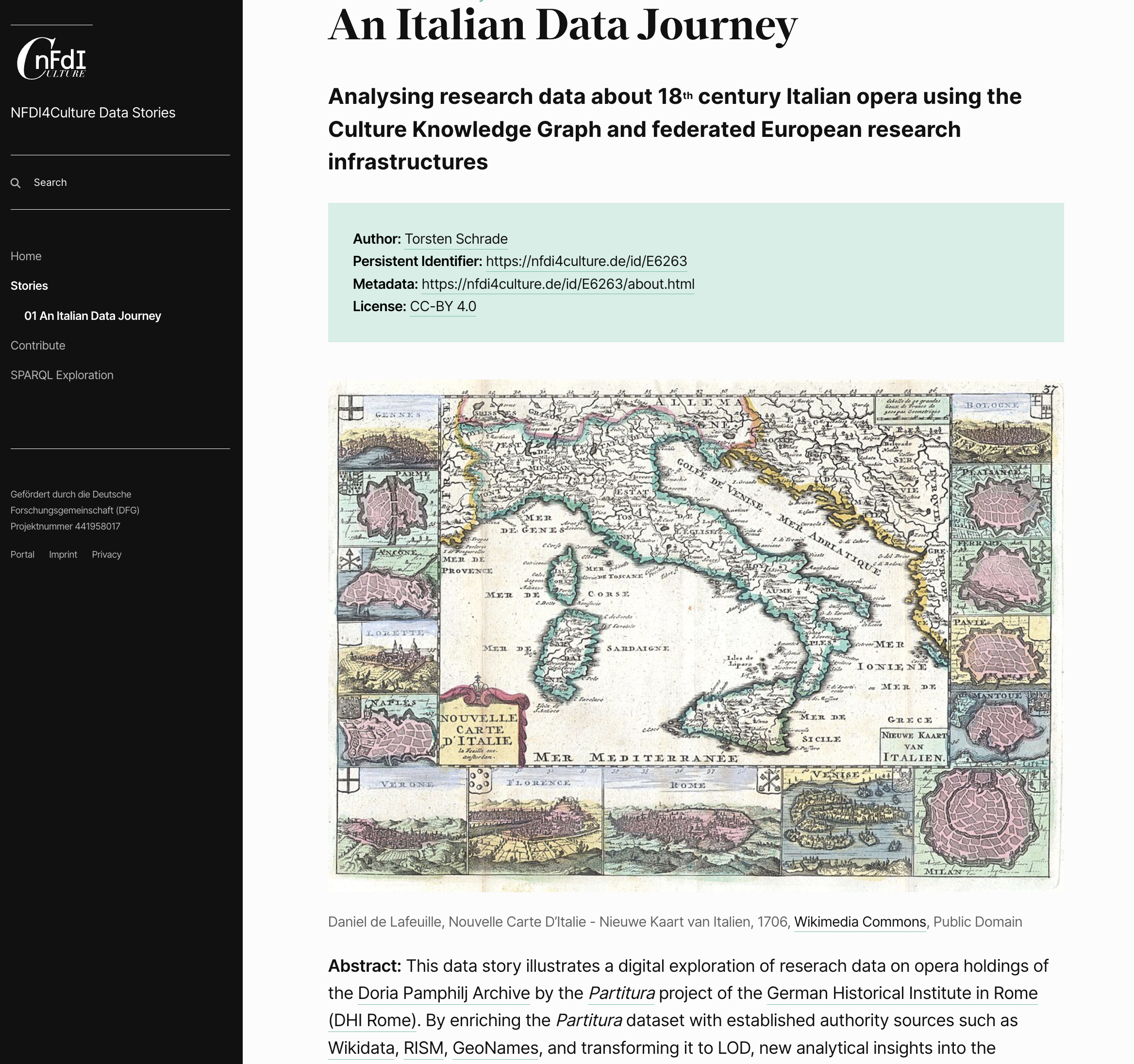}
    \caption{Data Story E6263 titled "An Italian Data Journey" by Torsten Schade}
    \label{fig:italianjourney}
\end{figure}

The graph’s discoverability is shaped by its modular design, the alignment of CTO with NFDIcore and BFO, and the deliberately light-weight ontology. Since it integrates multiple providers, its coverage remains uneven, with information density depending on each provider's specific domain. While external identifiers such as RISM\footnote{\url{https://rism.info/}} or GeoNames\footnote{\url{http://sws.geonames.org/}} and classification systems like Iconclass\footnote{\url{ https://iconclass.org/}} or Getty AAT\footnote{\url{https://www.getty.edu/research/tools/vocabularies/aat/}} facilitate data integration and querying, effective utilization requires knowledge of how they are represented within the underlying data. These challenges are further detailed in ~\cite{tietz2025knowledge}. Data stories help to address them by pairing specific questions with the queries that execute them over the graph. Embedding these queries within the narratives makes data conventions visible, demonstrates modeling patterns, illustrate what a given feed contains, and ensures that queries remain valid as the graph evolves.

One example of a data story based on the NFDI4Culture-KG is titled "An Italian Data Journey"\footnote{\url{https://datastories.nfdi4culture.de/story/E6263/}}, depicted in Fig.~\ref{fig:italianjourney}. It was authored by an expert in Digital Humanities and leverages services provided by NFDI4Culture and EOSC\footnote{\url{https://open-science-cloud.ec.europa.eu/}}. The story exemplifies how data federation with European infrastructures can significantly enhance the interoperability of research data and create multimodal research perspectives. Methodologically, it ranges from genre distribution analyses to geospatial mappings of opera premiere locations and music information retrieval through federated SPARQL queries. For a reader, its value is twofold. The story opens with a simple overview query that lists the data portals feeding the graph and the number of records of each type it holds, allowing new users to immediately understand the scale and composition of the available data. The subsequent analyses serve as reusable patterns: each embedded query can be opened, inspected, and adapted ensuring readers gain both direct answers and templates for their own questions.

Another example titled "A Deep Dive into NFDI4Culture's Integration Workflows"\footnote{\url{https://datastories.nfdi4culture.de/story/E6477/}} explores the connections between RISM Online Musical Sources\footnote{\url{https://rism.online/}} (an access point to over 1m musical sources of the Répertoire International des Sources Musicales) and the Gregorovius letters\footnote{\url{https://gregorovius-edition.dhi-roma.it/}} through the lens of the NFDI4Culture-KG. The Gregorovius edition contains 1,093 annotated pieces of correspondence from the historian Ferdinand Gregorovius (1821-1881), whose heritage from the 19th century testifies to a rich engagement with intellectual-historical movements and musicians of his time. 

This data story, further described in~\cite{soehn2025datastory}, demonstrates with query examples the way heterogeneous data are integrated in the KG and how external identifiers help to link once entirely separate feeds in a way that they can be queried through a single access point. 

Both story examples illustrate which questions the graph can answer and how to construct them. Furthermore, they demonstrate how heterogeneous data is integrated and connected in the first place. However, consuming a data story remains guided by the author's own narrative. To answer their own research questions, users must turn from consumer to author by creating their own data stories. 

\subsection{Authoring Data Stories: A SHMARQL-based Environment}

\begin{figure}
    \centering
    \includegraphics[width=1\linewidth]{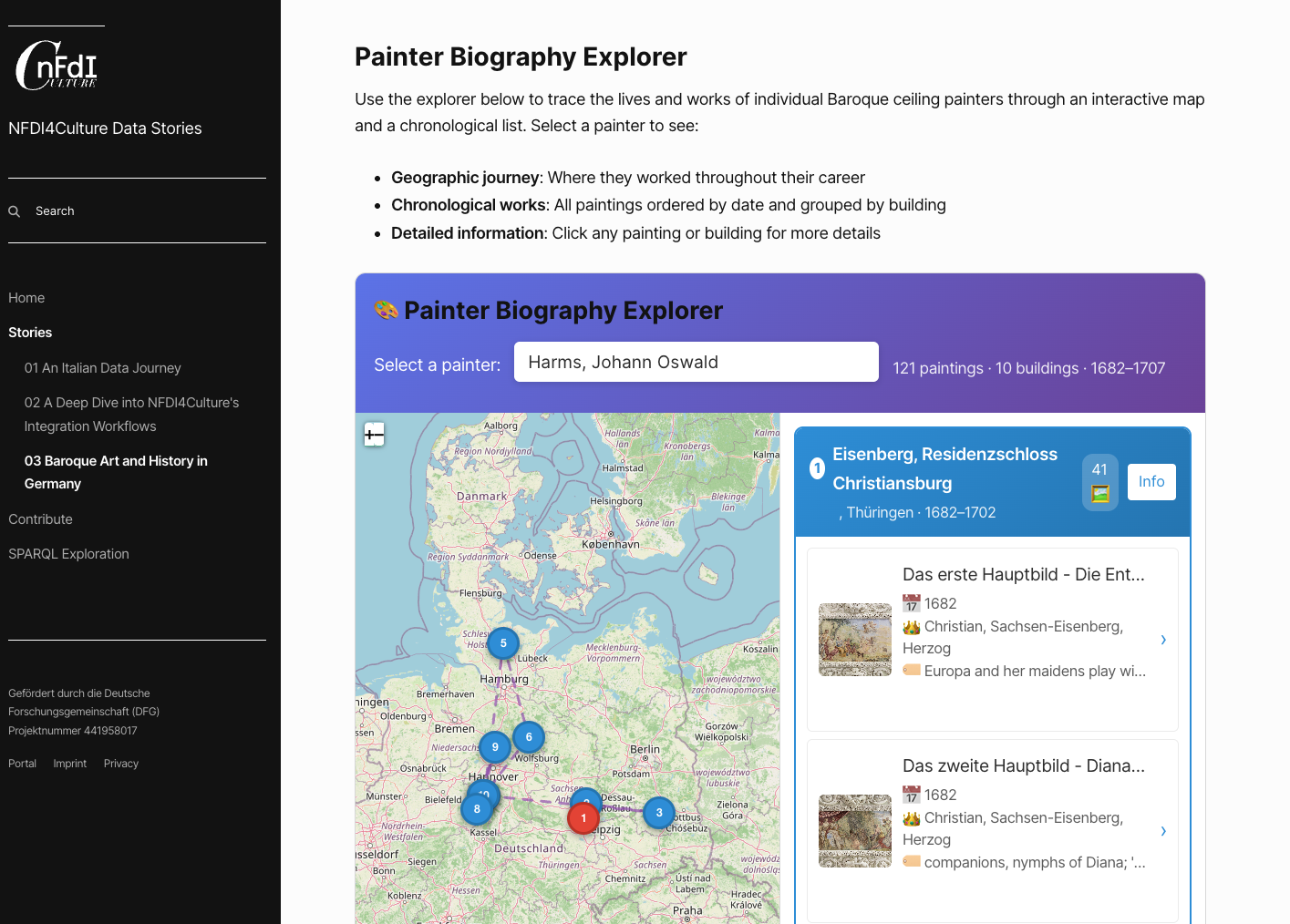}
    \caption{Explorer to trace the lives and works of individual Baroque ceiling painters through an interactive map, as part of the data story "Baroque Ceiling Paintings in Germany" by Thanos Drossos, Robin Kleemann, YiMin Cai.}
    \label{fig:cbdd-map}
\end{figure}

The SHMARQL platform enables the creation and publication of Linked Data applications by combining an RDF triplestore with Markdown-based querying and visualization features. Users can ingest raw triple files into Oxigraph\footnote{\url{https://github.com/oxigraph/oxigraph}}, an integrated high-performance triplestore for medium-sized KGs, or connect any SPARQL-compliant HTTP endpoint such as QLever~\cite{bast2025sparqloscope}, for larger KGs. By adding a custom Markdown code-block extension, SPARQL queries are executed directly, displaying their end-results within the narrative. In early versions of SHMARQL, configuration parameters for the Plotly\footnote{\url{https://plotly.com/python/}} visualization library were embedded within SPARQL query comments. This combined each query and its visualization setup into a single, portable and cacheable component. Rather than serving static graphs or tables, the system keeps these visualizations tied to live, editable queries. Readers can access and modify the underlying queries for further exploration or to inspect data provenance. Ultimately, a defining strength of this environment is that every claim in a story is backed by an explicit, inspectable query. 

\subsubsection{Lessons Learned on SHMARQL-supported Story Creation}

Insights into how the first SHMARQL-based authoring environment was received in practice came from a hands-on Master's seminar at Karlsruhe Institute of Technology (KIT) over several semesters. Students at KIT were provided with the NFDI4Culture-KG as a fixed resource for their work. The data story "Baroque Ceiling Paintings in Germany"\footnote{\url{https://datastories.ise.fiz-karlsruhe.de/story/CbDD/}} is one example of a result of the course and is depicted in Fig.~\ref{fig:cbdd-map}. The students independently selected their target data, research questions and the overall narrative. Coming from backgrounds in computer science and economics, most had no prior experience with digital humanities or the domain-specific research questions typical of this discipline. Students had varying levels of prior experience in semantic web technologies ranging from proficient users to complete beginners.

The authoring process proved to be an effective method for students to learn the graph's modeling patterns. By providing a self-contained story with a defined research goal and a fixed set of questions, the task gave data exploration a clear purpose, demonstrating far greater effectiveness than confronting participants with the open SPARQL endpoint alone. In pursuing their goal, the students became familiar with the data feeds and their underlying structures, and after initial onboarding, successfully learned to query and visualize the data. Participants responded positively to several key design choices. In particular, they  appreciated the seamless integration of queries with their corresponding result integrated directly within the narrative, as well as the ability to embed Plotly visualizations alongside the text. Furthermore, the use of Markdown was widely valued for keeping the authoring process lightweight and intuitive. However, writing SPARQL queries from scratch remained a considerable hurdle for participants without prior experience. As expected, the initial SHMARQL-based environment alone could not fully resolve this challenge. This limitation directly motivated the development of an environment that provides more active query construction support, as described in the following section.


\subsection{Towards AI Assisted Authoring with LODEON}

\begin{figure}
    \centering
    \includegraphics[width=1\linewidth]{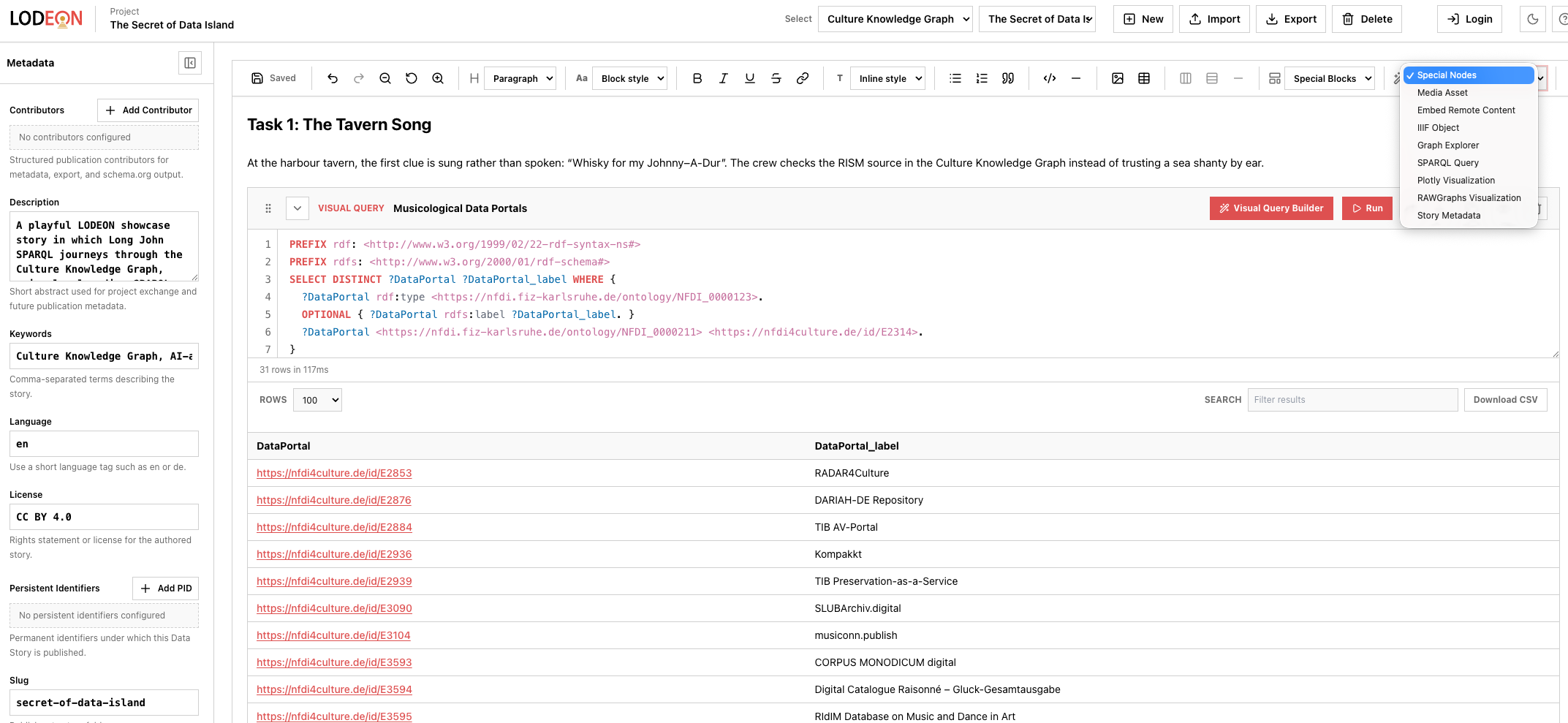}
    \caption{The LODEON authoring environment on the example of the story "Long John SPARQL and the Secret of Data Island" by Torsten Schrade}
    \label{fig:lodeon-overview}
\end{figure}

\begin{figure}
    \centering
    \begin{subfigure}[t]{0.55\linewidth}
        \centering
        \includegraphics[width=\linewidth]{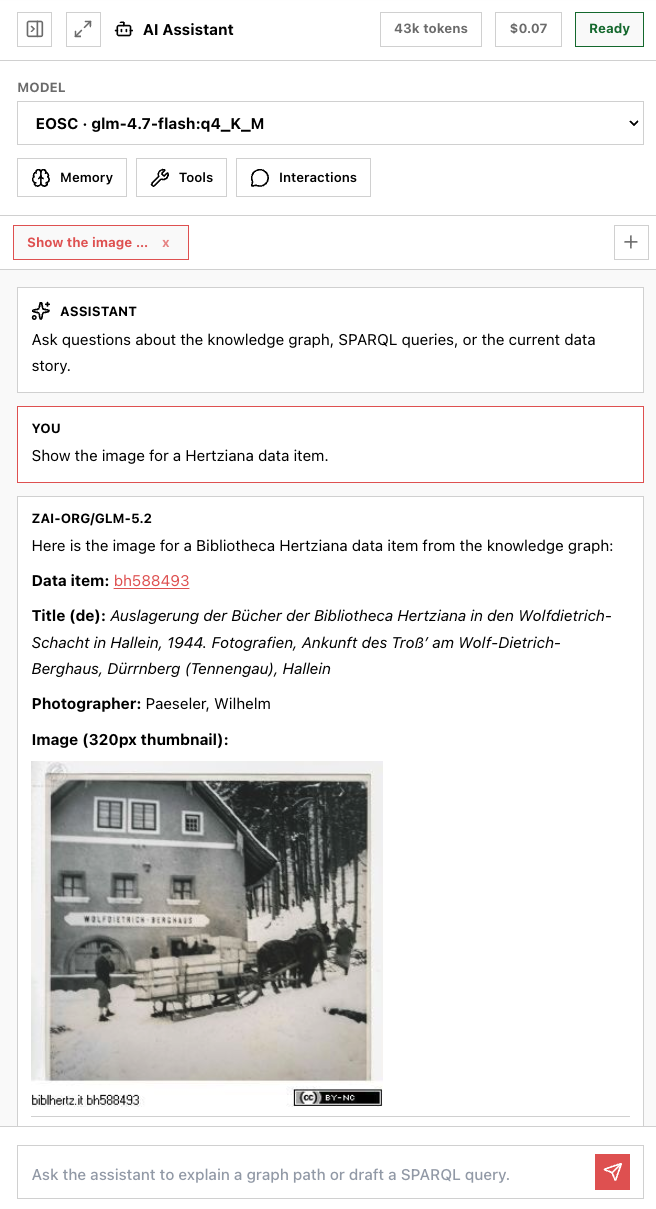}
        \caption{LODEON AI assistant for KG exploration and data story authoring.}
        \label{fig:AI}
    \end{subfigure}
    \hfill
    \begin{subfigure}[t]{0.41\linewidth}
        \centering
        \includegraphics[width=\linewidth]{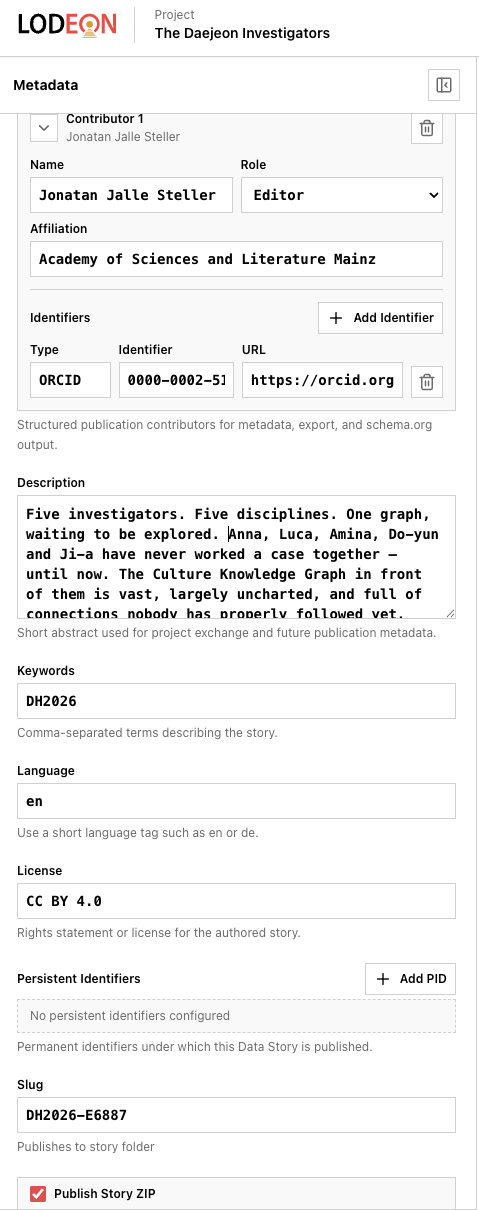}
        \caption{LODEON metadata panel.}
        \label{fig:metadata}
    \end{subfigure}
    \caption{Close-ups of LODEON authoring workbench: (a) the AI assistant and (b) the metadata panel.}
    \label{fig:lodeon-panels}
\end{figure}

LODEON\footnote{\url{https://adwmainz.pages.gitlab.rlp.net/digicademy/lodeon/lodeon-documentation/}} integrates KGs, WYSIWYG note-taking, evidence-based exploration, and guided AI assistance into a unified workbench. This enables users to interact with KGs, document their insights, and produce media-rich visualizations (cf. Fig.~\ref{fig:lodeon-overview}). Its local-first, data-sovereign architecture allows users to keep full control over their data and model choices on their own infrastructure. Currently, the public workbench supports the utilization of both the NFDI4Culture-KG and GeMeA, a KG containing data from the German Digital Library (DDB)\footnote{\url{https://github.com/ISE-FIZKarlsruhe/gemea}}. LODEON facilitates an evidence-based KG workflow. Authors can select an active KG, formulate questions, gather evidence, and synthesize findings into structured notes or narratives. SHMARQL remains part of LODEON and provides the underlying querying and rendering of results against the live endpoint. Query construction is further supported both by the visual query builder Sparnatural and by an AI assistant that offers graph-oriented help, depicted in Fig.~\ref{fig:AI}. The AI assistant is built on a Model Context Protocol (MCP)\footnote{\url{https://modelcontextprotocol.io}} shaped tool layer that makes its assistance traceable. The assistant can inspect the profile of the active KG, look up ontology terms, validate and run queries. It then returns proposals that the author can insert directly into the story. On this basis it can generate SPARQL queries, suggest how a story might continue, and create visualizations. To lower the initial barrier, the assistant starts from a set of curated query suggestions. In the local version of LODEON, users can connect their own model that exposes an OpenAI-compatible API.

Query results and visualizations can be inserted into the story as interactive nodes. Beyond Plotly charts, a story can incorporate a range of specialized nodes, including IIIF images, RAWGraphs visualisations, and a node to embed remote content, such as videos or interactive 3D models, as depicted in Fig.~\ref{fig:3D} in Appendix~\ref{ap:A}.  

LODEON also provides a metadata panel, shown in Fig.~\ref{fig:metadata}. The author can add a short abstract, keywords, a language tag, name contributors together with their roles, affiliations, and identifiers such as ORCID. Furthermore, the user can assign a license and one or more persistent identifiers under which the story will be published. This information is carried into the exported story, so that a published data story can be referenced and reused like any other research output. Once a story is complete, it can be previewed in the way it will be rendered and then exported as a zip file or published. Throughout this process, LODEON assists, suggests, queries, and visualizes, while the author remains in full control of what enters the story in the end.

\subsubsection{Availability of LODEON}
A demo video of the LODEON proof-of-concept is available on the Web\footnote{\url{https://drive.google.com/file/d/1eIPn1hrUoVWU4JccdI3L1HbVuF_A3fv-/view?usp=sharing}}, and the public workbench can be used online\footnote{\url{https://datastories.nfdi4culture.de/lodeon/}}. Due to the associated costs, LODEON's AI assistant is not yet freely available on the public workbench, but users are free to run LODEON locally and connect their own models, as it is available as MIT licensed open source software on GitLab~\footnote{\url{https://gitlab.rlp.net/adwmainz/digicademy/lodeon/}}.

\subsubsection{Lessons Learned on LODEON and AI Assisted Data Exploration}

LODEON was presented and tested as a proof-of-concept in a hands-on workshop at DH2026 in Daejeon, South Korea~\cite{schrade2026-lodeon}\footnote{\url{https://dh2026.adho.org/}}. Participants worked with a pre-structured data story and completed tasks involving the AI assistant, adding SPARQL queries and their results, building queries with Sparnatural, and creating visualizations. For this workshop, users received a login in order to use the AI assistant and could choose between two models, Qwen 3.5 and EOSC GLM 4.7 Flash. Feedback was gathered through live polls using the Claper tool\footnote{\url{https://claper.co/}} and largely through verbal discussion, since only about half of the roughly 60 participants joined the Claper event. 

The audience came from disciplines across languages and literature, digital humanities, history and archaeology, the arts, and computer science. Most participants had prior exposure to KGs, worked with graph databases, or integrated a KG into an application. A minority described themselves as comfortable with SPARQL. Data stories were largely new to the group, and most of those who participated in the poll had neither read nor written one before. The reception of LODEON in general was widely positive. The participants judged LODEON useful for their own work within cultural heritage and digital humanities and beyond. No participant judged the tool as irrelevant or unusable. The AI assistant in particular received a more mixed assessment. On a scale from 1 (not at all) to 10 (perfect), ratings of its results clustered in the lower range, mostly between 2 and 4. 

Graphical user interfaces and data stories alike are always subject to design decisions that predetermine what users are asked to do. These decisions include available filters, the order in which information is displayed or a pre-curated question of the AI assistant a data story author can choose. These design decisions quietly shape the space of possible questions and therefore may impact the research results. Direct access to the data, e.g. by means of a SPARQL query, is the counterweight, as it allows users to pursue their own interests and lines of research. One question that motivated the creation of LODEON was whether AI could bridge the two, giving users direct access to the data while also helping them understand what can be searched for. Experience now suggests that it can do so partially and under certain conditions. In the workshop, the AI assistant lowered the barrier to query construction and helped participants understand a graph they did not know beforehand. However, it sometimes also produced unreliable answers, particularly for aggregate questions such as counts, but its output looked equally trustworthy whether it was correct or not. In the aftermath of the workshop, much care has been taken to improve the quality and visual transparency of the AI assistant. The infrastructural conditions under which AI is used in research add to this, since availability is also determined by costs. On the public workbench the assistant cannot be offered freely yet and during the workshop, results had to be capped for efficiency. That means, when using the public workbench, performance and costs currently also determine what users see and what they do not. This is another reason why authors are advised to run their own models locally instead.  

These conditions do not argue against AI as a bridge, but there are risks and limits. The principle that every claim in a story remains backed by an explicit, inspectable query is necessary and relevant. However, the workshop also made clear that it is not sufficient on its own. Transparency has to further improve and it has to be made clear where a result may be wrong and where an answer also reflects the limits of the infrastructure and not only the data. Only then can AI provide both access and understanding without quietly reintroducing the design decisions that direct access to the data was meant to escape.

\section{Quality Assessment}
\label{sec:quality}

The quality of KGs and Linked Data has been studied extensively. For instance, Färber et al. assess DBpedia, Freebase, OpenCyc, Wikidata, and YAGO against a common set of data-quality criteria, such as accuracy, consistency, and completeness, applied through metrics computed over a graph as a whole~\cite{farber2017linked}. Such aggregate measures are effective at scale, but they can hide localized problems in individual data feeds, which might originate in the original data, the data integration step, or could reveal additional requirements for the ontology. Authoring a data story brings such problems to light, since concrete research questions require records to be retrieved and inspected in depth. In this way, issues such as missing entities, or incomplete metadata become visible in a way that graph-wide statistics do not reveal. Story authoring is therefore better understood as a complement to automated quality assessment than as a replacement for it.

In the process of authoring a data story during the seminar at KIT\footnote{\url{https://datastories.ise.fiz-karlsruhe.de/story/CbDD/}}, the Master's students were not only able to test the SHMARQL-based authoring environment, but also uncovered a range of errors in the graph. The process of writing a narrative helped in discovering missing data points, and missing controlled vocabulary entries to enable standardized data linking. The full report is available on Github\footnote{\url{https://github.com/ThanosDrossos/course-data-stories-baroque/blob/main/src/docs/story/CbDD/nfdi4culture-kg-issues.md}}. As a consequence, several of the missing fields could be added to the NFDI4Culture-KG in subsequent ingestion rounds, and missing controlled-vocabulary entries needed for standardized linking could be identified. Currently, users can only report on issues discovered in the KG while authoring a data story by means of Github issues or by contacting NFDI4Culture. However, in future development, a dedicated and more structured means of data quality issue reporting with respect to the KG is planned. This also involves the report of missing connections, which may reveal new requirements for the underlying ontology, as more and more users contribute their stories and their research questions.



\section{Conclusion}
\label{sec:conclusion}

This paper presents data stories as a way to make the large, heterogeneous, and continuously growing NFDI4Culture-KG more accessible, and to turn its exploration into a form of quality assessment. The paper follows data stories in NFDI4Culture from reading to authoring. Existing stories, such as \textit{An Italian Data Journey} and \textit{A Deep Dive into NFDI4Culture's Integration Workflows}, give new users an entry point to the graph and a set of reusable query patterns. A first authoring environment built with SHMARQL has been extended by the LODEON workbench, a proof-of-concept that adds Sparnatural and an AI assistant. 

Early experience, including a workshop at DH2026 suggests that this approach is useful. The reception of LODEON is widely positive, and the AI assistant lowers the barrier to query construction and helps participants understand an unfamiliar graph. At the same time, it became clear that the AI assistant can serve as a bridge to the data in a research process only when its own limits are made transparent. Immediate next steps include a quantitative evaluation of the AI assistant focusing on the accuracy of the SPARQL queries it generates, and on which model performs best for the specific task of data story generation at the lowest computational cost. LODEON will be further developed and improved based on the existing lessons learned and the upcoming quantitative results, before the current proof-of-concept can be officially released. The challenges addressed in this paper are not specific to the NFDI4Culture project or to the cultural heritage field. Therefore, future development will also include the use cases in domains, like Materials Science and Sports Science.

\begin{acknowledgments}
 This work is funded by Deutsche Forschungsgemeinschaft (DFG), project number 441958017. \\ 
 We would also like to thank YiMin Cai, Thanos Drossos, and Robin Kleemann who completed the Master's seminar "Knowledge-Driven AI" with the topic "Telling Data Stories with Semantic Technologies and Generative AI" at Karlruhe Institute of Technology (KIT). As part of the seminar, the students authored the data story "Baroque Ceiling Paintings in Germany" and provided useful insights into existing data quality challenges in the NFDI4Culture-KG. 
\end{acknowledgments}

\section*{Declaration on Generative AI}

 During the preparation of this work, the author(s) used Claude Opus 4.8 in order to: Grammar and spelling check, and paraphrase and reword. After using these tool(s)/service(s), the author(s) reviewed and edited the content as needed and take(s) full responsibility for the publication’s content. 


\bibliography{bibliography}

\appendix

\section{LODEON Features}
\label{ap:A}

\begin{figure}[h!] 
    \centering
    \includegraphics[width=1.0\linewidth]{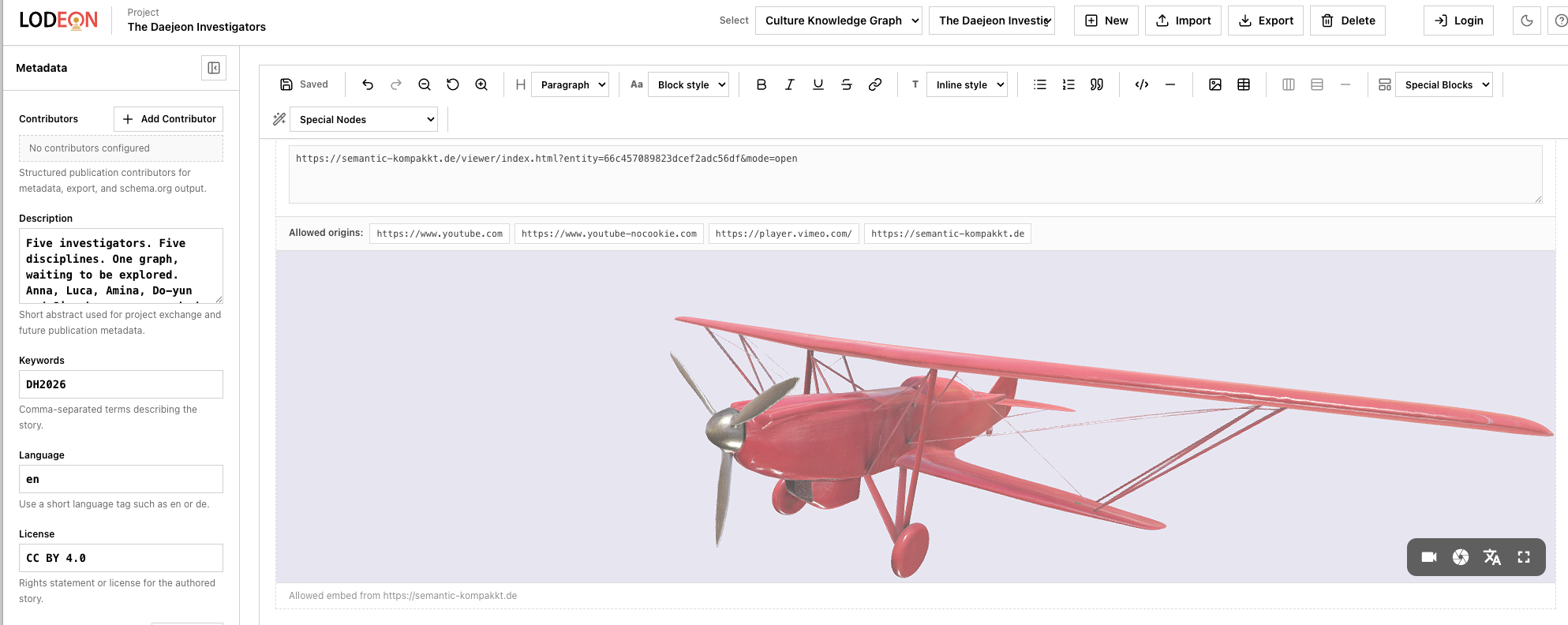}
    \caption{LODEON supports the embedding of remote content, like the interactive 3D model from the data portal semantic kompakkt (\url{https://nfdi4culture.de/id/E3752}). }
    \label{fig:3D}
\end{figure}

\end{document}